\documentclass[letterpaper, 10 pt, conference]{ieeeconf}  

 \usepackage{amsmath}
 \usepackage{graphicx}
 \usepackage{xcolor}
 \usepackage{booktabs}
\usepackage{graphicx}
\usepackage{makecell}
\usepackage{rotating}
\usepackage{threeparttable}
\usepackage{amssymb}
\usepackage{multirow}
\usepackage[table]{xcolor}
\definecolor{myLightGray}{HTML}{EFEFEF}

\title{\LARGE \bf
VLMFusionOcc3D: VLM Assisted Multi-Modal 3D Semantic Occupancy Prediction
}

\author{A. Enes Doruk and Hasan F. Ates
\thanks{The authors are with the Department of Artificial Intelligence and Data Engineering, Ozyegin University, Istanbul, Turkey. Emails: {\tt\small enes.doruk@ozyegin.edu.tr, hasan.ates@ozyegin.edu.tr}}
}

\begin{document}

\maketitle
\thispagestyle{empty}
\pagestyle{empty}

\begin{abstract}
This paper introduces VLMFusionOcc3D, a robust multimodal framework for dense 3D semantic occupancy prediction in autonomous driving. Current voxel-based occupancy models often struggle with semantic ambiguity in sparse geometric grids and performance degradation under adverse weather conditions. To address these challenges, we leverage the rich linguistic priors of Vision-Language Models (VLMs) to anchor ambiguous voxel features to stable semantic concepts. Our framework initiates with a dual-branch feature extraction pipeline that projects multi-view images and LiDAR point clouds into a unified voxel space. We propose Instance-driven VLM Attention (InstVLM), which utilizes gated cross-attention and LoRA-adapted CLIP embeddings to inject high-level semantic and geographic priors directly into the 3D voxels. Furthermore, we introduce Weather-Aware Adaptive Fusion (WeathFusion), a dynamic gating mechanism that utilizes vehicle metadata and weather-conditioned prompts to re-weight sensor contributions based on real-time environmental reliability. To ensure structural consistency, a Depth-Aware Geometric Alignment (DAGA) loss is employed to align dense camera-derived geometry with sparse, spatially accurate LiDAR returns. Extensive experiments on the nuScenes and SemanticKITTI datasets demonstrate that our plug-and-play modules consistently enhance the performance of state-of-the-art voxel-based baselines. Notably, our approach achieves significant improvements in challenging weather scenarios, offering a scalable and robust solution for complex urban navigation.
\end{abstract}

\section{Introduction}

Understanding the complex three-dimensional (3D) geometry and semantic distribution of a vehicle's surroundings is a cornerstone of safe autonomous driving (AV) \cite{caesar2020nuscenes, behley2019semantickitti}. While traditional 3D object detection provides bounding boxes for known categories, 3D Semantic Occupancy Prediction has emerged as a more comprehensive representation \cite{li2023bevformer, huang2023triocc}. By partitioning the environment into a dense voxel grid and assigning semantic labels to each cell, occupancy-based methods offer a geometrically consistent view of the world, effectively handling general obstacles and occluded regions that bounding-box-based methods might overlook \cite{yan2024inverse, gan2024coocc}.

Despite significant progress in multimodal fusion strategies, current state-of-the-art (SOTA) occupancy models face two critical hurdles. First, semantic ambiguity remains a persistent issue in raw voxel space \cite{zhu2024occmamba}. Geometric features alone are often insufficient to distinguish between morphologically similar classes, such as a pedestrian standing near a slender utility pole \cite{zhang2024mconet}. Second, environmental sensitivity poses a major challenge for reliability \cite{zhou2024occfusion}. Camera-based systems suffer from contrast loss in low-light conditions, while LiDAR sensors experience significant signal scattering during precipitation \cite{philion2020lift}. Existing fusion methods often employ static weightings that fail to adapt to these environmental degradations, leading to compromised perception in adverse weather.

\vspace{-0.6cm}
\begin{figure}[!ht]
  \centering
\includegraphics[width=0.8\columnwidth]{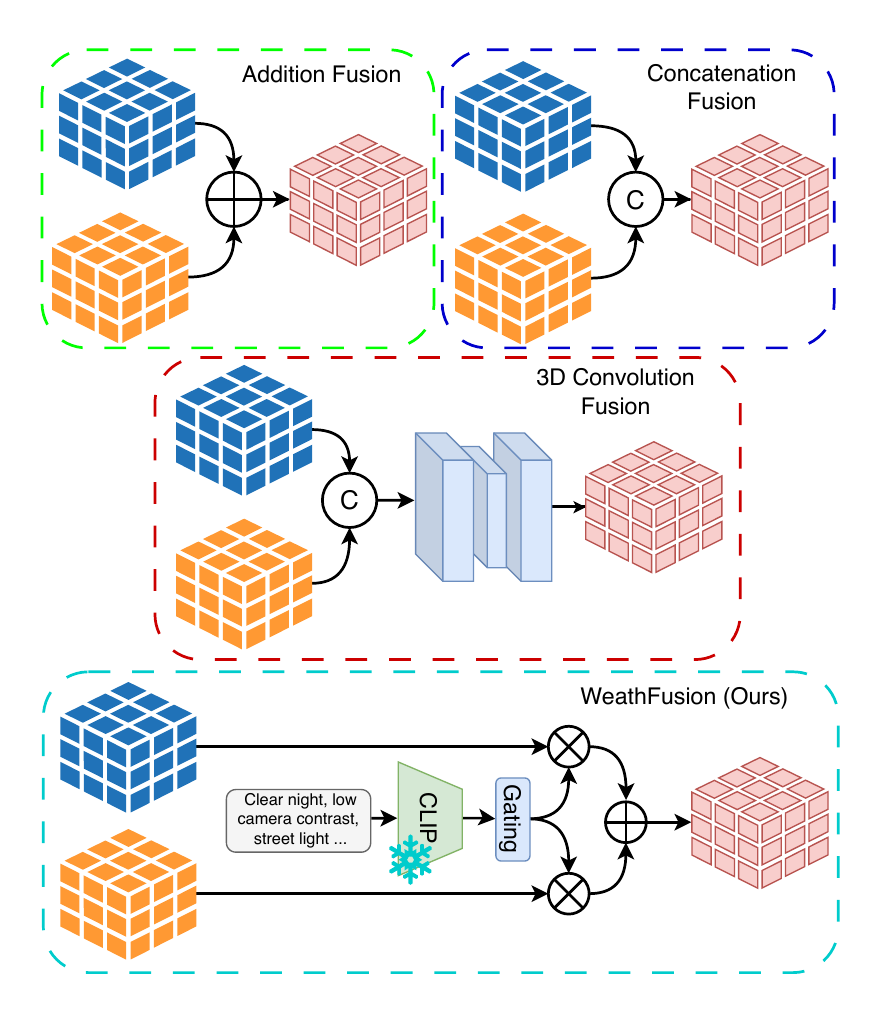}  
\vspace{-0.6cm}
\caption{Comparison of feature fusion architectures. Standard approaches include Addition Fusion for element-wise summation, Concatenation Fusion for channel-wise stacking, and 3D Convolution Fusion for learned spatial refinement. Our proposed WeathFusion introduces a language-aware gating mechanism. It utilizes a frozen CLIP encoder to process environmental descriptions, modulating input features through learned gating weights before final summation.}
  \label{fig:fusions}
\end{figure}

To address these limitations, we propose VLMFusionOcc3D, a multimodal 3D semantic occupancy framework that leverages the semantic common sense of Vision-Language Models (VLMs) and weather-aware context to achieve robust perception \cite{radford2021learning, wang2023bevclip}. We observe that VLMs, such as CLIP \cite{radford2021learning}, contain rich, category-agnostic linguistic priors that can anchor ambiguous 3D features to stable semantic concepts. Furthermore, we argue that weather-aware information provides an objective prior for environmental context that can be used to dynamically re-calibrate sensor trust.

\begin{figure*}[!ht]
  \centering
  \includegraphics[width=0.9\textwidth]{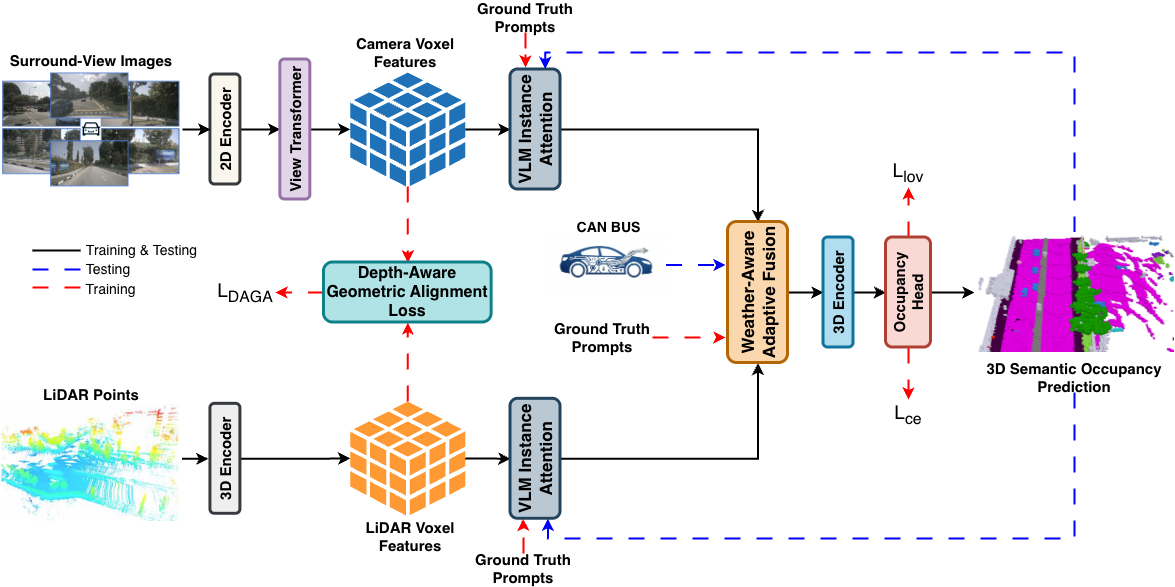}
  \vspace{-0.3cm}
  \caption{\textbf{Overall architecture of VLMFusionOcc3D.} Our pipeline integrates multi-view camera images and LiDAR point clouds through a dual-branch architecture. High-level semantic priors are injected via Instance-driven VLM Attention (InstVLM), followed by dynamic modality integration using the Weather-Aware Adaptive Fusion (WeathFusion) module. The final dense occupancy grid is optimized using the Depth-Aware Geometric Alignment (DAGA) loss.}
  \label{fig:overview}
\end{figure*}

The core of our framework consists of three novel components. First, we introduce Instance-driven VLM Attention (InstVLM), which utilizes gated cross-attention to inject high-level semantic and geographic priors into the voxel space \cite{hu2023lora}. Second, we present Weather-Aware Adaptive Fusion (WeathFusion), a dynamic gating system that utilizes weather-conditioned prompts and vehicle metadata to re-weight camera and LiDAR contributions based on current environmental conditions. Finally, to resolve the structural discrepancies between dense camera-derived voxel frustums and sparse LiDAR returns, we propose the Depth-Aware Geometric Alignment (DAGA) loss.

We evaluate our framework by integrating these modules into two high-performance voxel-based baselines, OccMamba \cite{zhu2024occmamba} and MCoNet \cite{zhang2024mconet}, demonstrating significant performance gains on the nuScenes \cite{caesar2020nuscenes} and SemanticKITTI \cite{behley2019semantickitti} benchmarks. Our main contributions are summarized as follows:

\begin{itemize}
\item We propose InstVLM, a parameter-efficient module that utilizes LoRA-adapted \cite{hu2023lora} VLM embeddings and gated cross-attention to resolve semantic ambiguity in 3D voxel grids.
\item We introduce WeathFusion, an adaptive fusion mechanism that dynamically modulates modality weights based on real-time weather-aware context derived from vehicle metadata.
\item We design the DAGA loss, which employs depth-dependent weighting and vertical sharpness constraints to align camera-derived geometry with spatially accurate LiDAR features.
\item Extensive experiments demonstrate that our plug-and-play modules consistently improve the mIoU of baseline architectures, particularly in challenging weather conditions.
\end{itemize}

\section{Related Work}
\subsection{Multimodal 3D Semantic Occupancy Prediction}
Current research emphasizes the integration of multi-view camera images with LiDAR point clouds to capitalize on high-resolution textures and precise ranging \cite{caesar2020nuscenes, behley2019semantickitti, philion2020lift}. Co-Occ pioneered this by coupling explicit LiDAR-camera feature fusion with implicit volume rendering regularization, effectively bridging the 3D-2D gap \cite{gan2024coocc}. To address the computational bottlenecks inherent in dense voxel grids, OccMamba introduced State Space Models (SSMs) to the 3D domain, achieving linear computation complexity while maintaining a global receptive field \cite{zhu2024occmamba}. Similarly, FusionOcc focuses on mitigating domain-specific noise by fusing modalities in both 2D and 3D spaces, utilizing dense depth maps to improve the transition from image space to voxel space \cite{li2023bevformer, huang2023triocc}.

Recently, Gaussian-based models like GaussianFormer \cite{huang2025gaussianformer} and GaussianOcc3D \cite{doruk2026gaussianocc3d} have emerged, representing scenes as sparse 3D Gaussians for improved memory efficiency. However, these primitive-based methods often lack the inherent neighborhood consistency and structured spatial priors found in regular voxel grids, which can lead to geometric artifacts in sparse regions. Our framework maintains a voxel-based representation to ensure unambiguous spatial coverage and superior semantic-geometric consistency.

Robustness in varying conditions has also been addressed through expanded sensor suites. OccFusion \cite{zhou2024occfusion} integrates surround-view cameras, LiDAR, and radar features to maintain perception reliability during sensor degradation \cite{zhou2024occfusion}. Concurrently, MCoNet \cite{zhang2024mconet} leverages multi-view temporal consistency and stereo-matching priors to resolve the ill-posed nature of depth estimation \cite{zhang2024mconet}, while M3Net \cite{chen2025m3net} adopts a multi-task learning approach to synchronize occupancy prediction with detection and segmentation. Recent advancements like InverseMatrixVT3D \cite{yan2024inverse} have further refined these representations using joint semantic-depth guidance and efficient projection matrices \cite{yan2024inverse}. Our proposed framework builds upon these foundations by introducing a weather-aware gating system that dynamically re-weights sensor contributions based on objective vehicle metadata.

\subsection{VLMs in 3D Semantic Occupancy Prediction}
The integration of Vision-Language Models (VLMs) such as CLIP \cite{li2023bevclip} has emerged as a key solution for addressing long-tail class detection and semantic sparsity. Pop-Occ \cite{vobecky2023pop3d}  utilizes text-aligned features to enable zero-shot 3D grounding and semantic labeling without predefined closed-set vocabularies. Building on multimodal alignment, BEV-CLIP \cite{li2023bevclip} utilizes descriptive text to represent complex scenes in BEV space, leveraging LLMs to enhance semantic richness. Recently, Language Driven Occupancy Prediction (LOcc \cite{yu2024locc}) introduced a semantic transitive labeling pipeline to generate dense language occupancy ground truth, transferring labels from images to LiDAR and ultimately to voxels.

To improve efficiency, VLM-Occ \cite{sun2023vlmocc} explores the use of frozen VLM backbones for zero-shot occupancy, whereas PromptOcc \cite{wu2024promptocc} introduces learnable prompts to better align 3D features. Our approach differentiates itself by adapting a Gated Cross-Attention mechanism specifically for 3D voxel spaces. Unlike current methods that apply a global semantic prior, our model uses 3D geometric context to gate the VLM influence, ensuring that linguistic embeddings only refine high-relevance voxels.

\section{Methodology}
Our framework adopts a multimodal voxel-based architecture to generate a dense $3D$ semantic occupancy grid by integrating 6-view surround camera images, $\mathcal{I} = \{I_i\}_{i=1}^{6}$, and LiDAR point clouds $\mathcal{P}$. The pipeline initiates with a dual-branch feature extraction: the camera branch employs a ResNet-50 backbone with a Feature Pyramid Network (FPN) to project multi-scale features into a voxel space $V_{cam}$ via a Lift-Splat-Shoot (LSS)-based view transformer. Concurrently, the LiDAR branch processes $\mathcal{P}$ through a voxelization layer and a sparse 3D encoder to produce the geometric voxel volume $V_{pts}$.

The core contribution lies in the subsequent refinement and fusion, as illustrated in the Fig. \ref{fig:overview}. We first introduce Instance-driven VLM Attention (InstVLM), which utilizes gated cross-attention independently on both branches to inject high-level semantic priors. These mechanisms are guided by semantic and environmental context prompts processed through a CLIP encoder specialized via Low-Rank Adaptation (LoRA). Following this, Weather-Aware Adaptive Fusion (WeathFusion) employs gated attention to achieve dynamic modality integration based on real-time environmental conditions. The unified volume is then processed through a 3D ResNet-FPN encoder to extract deep spatial-semantic correlations for the occupancy head. Finally, the Depth-Aware Geometric Alignment (DAGA) loss is applied to resolve the structural discrepancies between the dense camera-derived frustums and sparse LiDAR returns.

\begin{figure}[!ht]
  \centering
  \includegraphics[width=\columnwidth]{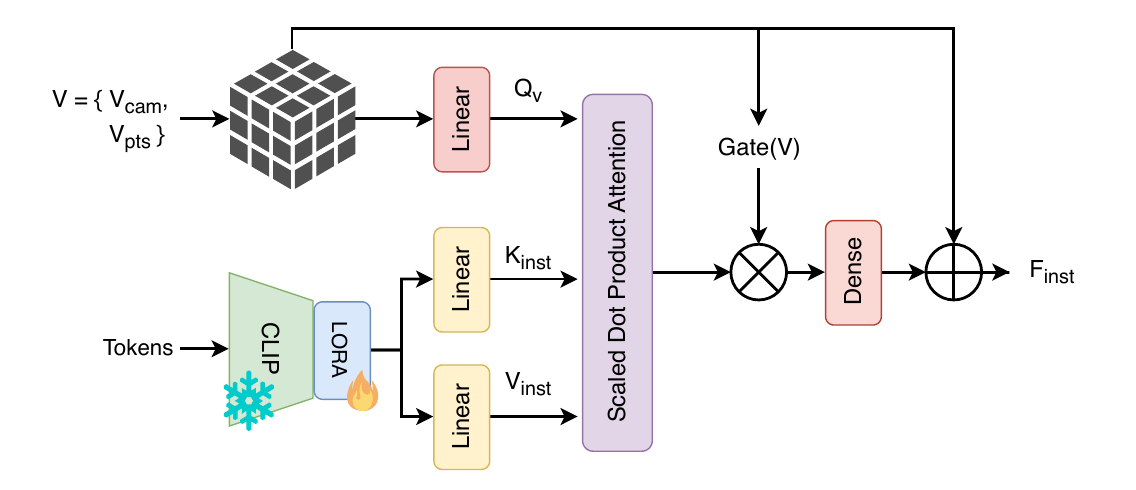}
  \vspace{-0.7cm}
  \caption{Architecture of the Instance-driven VLM Attention (InstVLM) module. The module employs a gated cross-attention mechanism to anchor 3D voxel features to continuous text embeddings from a LoRA-adapted CLIP encoder. The gating mechanism ensures that semantic information is selectively fused into spatial voxels.}
  \label{fig:instvlm}
\end{figure}

\subsection{Instance-driven VLM Attention}
Semantic ambiguity in 3D voxel grids often impedes precise occupancy prediction. Even though we construct our voxel representation from both multi-view visual features and LiDAR point clouds—resulting in a semantic-aware voxel space—these features alone may lack the categorical cues required to distinguish morphologically similar classes, such as pedestrians versus slender poles. We address this by leveraging the pre-trained latent space of a Vision-Language Model (VLM) to anchor these reconstructed, semantic-aware voxel features to continuous text embeddings, ensuring that the structural data is reinforced by robust linguistic priors.

\vspace{-0.7cm}
\begin{figure}[!ht]
  \centering \includegraphics[width=0.8\columnwidth]{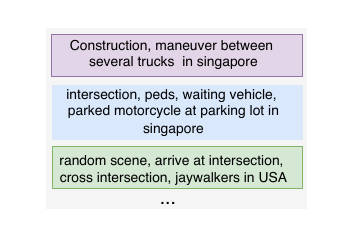}
\vspace{-1cm}
\caption{Examples of structured instance prompts. These prompts encapsulate category-specific information and geographic context to provide the VLM with rich environmental priors. During inference, a recursive strategy is employed to maintain temporal semantic stability.}
  \label{fig:instprompts}
\end{figure}

This process begins with generating structured prompts encapsulating class information and geographic context (e.g., USA vs. Singapore), which accounts for regional variances in vehicle morphology and urban layouts. The diversity and structure of these inputs are illustrated in Fig. \ref{fig:instprompts}. While ground-truth labels are used for prompting during training, we implement a recursive prompting strategy during inference to ensure temporal stability. A generic prompt is initialized at $t=0$, while for subsequent frames, the model's own semantic predictions from $t$ generate refined, category-specific prompts for $t+1$. Although predictions from the previous frame are utilized, this does not cause significant semantic drift during inference.To further ground these prompts, we integrate static geographic context provided via the CAN BUS, such as the specific driving region (e.g., USA or Singapore), which acts as a high-level semantic prior for class distribution. These prompts are processed via a LoRA-adapted CLIP encoder to ensure parameter-efficient adaptation for real-time deployment.

To integrate these priors, we adapt a gated cross-attention mechanism for 3D voxel spaces, the architecture of which is detailed in the InstVLM module in Fig. \ref{fig:instvlm}. To manage the non-linearity of sparse 3D data and prevent attention-sink artifacts, we employ a gating mechanism adapted directly to the 3D voxel features, ensuring semantic information is only fused into high-relevance voxels:
\begin{equation}
    F_{inst} = \text{Softmax}\left(\frac{Q_V K_{inst}^\top}{\sqrt{d_k}}\right) V_{inst}  \text{Gate}(V) + V
\end{equation}
where $Q_V$ represents the 3D voxel queries from $V_{cam}$ or $V_{pts}$, and $\text{Gate}(V)$ modulates influence based on the 3D geometric context $V$. This gate is implemented via a linear transformation followed by a Sigmoid activation, mapping the voxel features to a normalized weight space to ensure that linguistic embeddings only refine high-relevance voxels.

\begin{figure}[!ht]
  \centering
\includegraphics[width=\columnwidth]{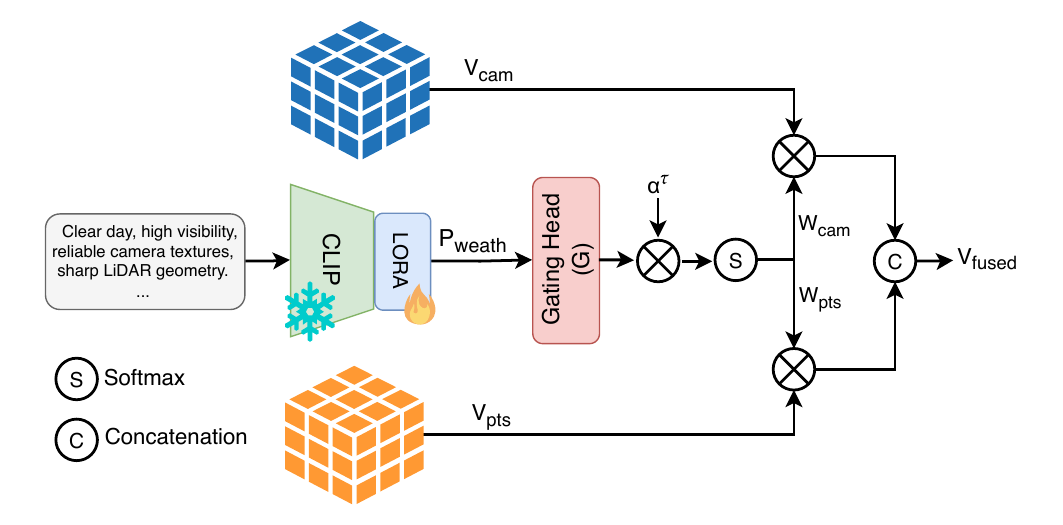}  
\vspace{-0.7cm}
\caption{Detailed schematic of the Weather-Aware Adaptive Fusion (WeathFusion) module. The gating head processes weather context prompts to compute dynamic reliability weights for each modality. This process enables the framework to robustly transition between sensors in response to environmental degradation, such as LiDAR scattering or low-light camera noise.}
  \label{fig:weathfusion}
\end{figure}

\subsection{Weather-Aware Adaptive Fusion}
To mitigate performance degradation in adverse environments, the WeathFusion module dynamically modulates sensor contributions based on real-time reliability. The structural implementation of this module is illustrated in Fig. \ref{fig:weathfusion}. Physical limitations—specifically LiDAR scattering in precipitation and camera contrast loss in low-light environments—necessitate a strategy that can transition away from corrupted data streams.

\begin{figure}[!ht]
  \centering
  \includegraphics[width=0.8\columnwidth]{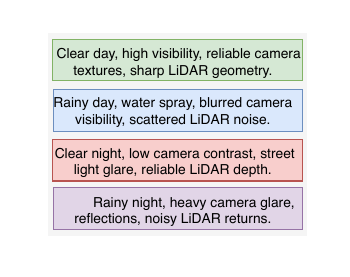}
  \vspace{-1cm}
  \caption{Environmental context prompts for weather adaptation. These prompts represent various operational domains. These embeddings supervise the gating head to prioritize robust sensors, such as favoring LiDAR in low-visibility or Cameras in clear conditions.}
  \label{fig:weatherprompts}
\end{figure}

During training, we utilize ground-truth weather labels to supervise the mapping between conditions and sensor trust. Examples of the structured weather prompts used to condition this fusion are provided in Fig. \ref{fig:weatherprompts}. For inference, ground-truth labels are replaced by metadata from the vehicle's CAN BUS. This telemetry-based approach is motivated by the relative stability of weather conditions over short intervals, providing a reliable environmental prior without the latency of an integrated prediction sub-network. The CLIP embedding informs a gating head—consisting of a two-layer MLP with a 32-dimensional hidden space and ReLU activation—to compute sensor logits and a learnable dynamic scale factor $\alpha^{\tau}$, acting as an inverse temperature to control weight distribution sharpness:
\begin{equation}
    \{w_{cam}, w_{pts}\} = \text{Softmax}\left(G(P_{weath})  \alpha^{\tau}\right)
\end{equation}
The weights scale the refined features to produce the fused volume $V_{fused}$:
\begin{equation}
    V_{fused} = \text{Concat}\left(V_{cam} w_{cam}, V_{pts} w_{pts}\right)
\end{equation}
This allows the framework to robustly prioritize the most reliable modality, ensuring consistent navigation across operational domains.

\definecolor{barriercolor}{RGB}{255, 0, 0}
\definecolor{bicyclecolor}{RGB}{0, 255, 0}
\definecolor{buscolor}{RGB}{255, 255, 0}
\definecolor{carcolor}{RGB}{0, 0, 255}
\definecolor{construction_vehiclecolor}{RGB}{255, 165, 0}
\definecolor{motorcyclecolor}{RGB}{128, 0, 128}
\definecolor{pedestriancolor}{RGB}{255, 192, 203}
\definecolor{traffic_conecolor}{RGB}{255, 69, 0}
\definecolor{trailercolor}{RGB}{192, 192, 192}
\definecolor{truckcolor}{RGB}{139 ,69 ,19}
\definecolor{driveable_surfacecolor}{RGB}{135 ,206 ,235}
\definecolor{other_flatcolor}{RGB}{160 ,82 ,45}
\definecolor{sidewalkcolor}{RGB}{211 ,211 ,211}
\definecolor{terraincolor}{RGB}{139 ,105 ,20}
\definecolor{manmadecolor}{RGB}{112 ,128 ,144}
\definecolor{vegetationcolor}{RGB}{34 ,139 ,34}

\newcommand{\makecolorcell}[2]{%
    \makecell[b]{\begin{turn}{90}\colorbox{#1}{\hspace{3pt}\rule{0pt}{3pt}} #2\end{turn}}}

\begin{table*}[t]
\centering
\caption{Quantitative comparisons on the nuScenes-OpenOccupancy validation set. C, D, L denote camera, depth and LiDAR, respectively. The best and second-best are in bold and underlined, respectively.}
\vspace{-0.3cm}
\resizebox{0.9\linewidth}{!}{
\setlength{\tabcolsep}{1mm}
\begin{threeparttable}
\begin{tabular}{l|c|cc|cccccccccccccccc}
\toprule

\makecell[b]{Method} & \makecell[b]{Input\\Modality} & 
\makecell[b]{IoU} & \makecell[b]{mIoU} & 
\makecolorcell{barriercolor}{barrier} & 
\makecolorcell{bicyclecolor}{bicycle} & 
\makecolorcell{buscolor}{bus} & 
\makecolorcell{carcolor}{car} & 
\makecolorcell{construction_vehiclecolor}{const. veh.} & 
\makecolorcell{motorcyclecolor}{motorcycle} & 
\makecolorcell{pedestriancolor}{pedestrian} & 
\makecolorcell{traffic_conecolor}{traffic cone} & 
\makecolorcell{trailercolor}{trailer} & 
\makecolorcell{truckcolor}{truck} & 
\makecolorcell{driveable_surfacecolor}{drive surf.} & 
\makecolorcell{other_flatcolor}{other\_flat} & 
\makecolorcell{sidewalkcolor}{sidewalk} & 
\makecolorcell{terraincolor}{terrain} & 
\makecolorcell{manmadecolor}{manmade} & 
\makecolorcell{vegetationcolor}{vegetation} \\

\midrule

MonoScene~\cite{cao2022monoscene} & C & 18.4 & 6.9 & 7.1 & 3.9 & 9.3 & 7.2 & 5.6 & 3.0 & 5.9 & 4.4 & 4.9 & 4.2 & 14.9 & 6.3 & 7.9 & 7.4 & 10.0 & 7.6 \\
TPVFormer~\cite{huang2023tri} & C & 15.3 & 7.8 & 9.3 & 4.1 & 11.3 & 10.1 & 5.2 & 4.3 & 5.9 & 5.3 & 6.8 & 6.5 & 13.6 & 9.0 & 8.3 & 8.0 & 9.2 & 8.2 \\
SparseOcc~\cite{tang2024sparseocc} & C & 21.8 & 14.1 & 16.1 & 9.3 & 15.1 & 18.6 & 7.3 & 9.4 & 11.2 & 9.4 & 7.2 & 13.0 & 31.8 & 21.7 & 20.7 & 18.8 & 6.1 & 10.6 \\
3DSketch~\cite{chen20203d} & C\&D & 25.6 & 10.7 & 12.0 & 5.1 & 10.7 & 12.4 & 6.5 & 4.0 & 5.0 & 6.3 & 8.0 & 7.2 & 21.8 & 14.8 & 13.0 & 11.8 & 12.0 & 21.2 \\
AICNet~\cite{li2020anisotropic} & C\&D & 23.8 & 10.6 & 11.5 & 4.0 & 11.8 & 12.3 & 5.1 & 3.8 & 6.2 & 6.0 & 8.2 & 7.5 & 24.1 & 13.0 & 12.8 & 11.5 & 11.6 & 20.2 \\
LMSCNet~\cite{roldao2020lmscnet} & L & 27.3 & 11.5 & 12.4 & 4.2 & 12.8 & 12.1 & 6.2 & 4.7 & 6.2 & 6.3 & 8.8 & 7.2 & 24.2 & 12.3 & 16.6 & 14.1 & 13.9 & 22.2 \\
JS3C-Net~\cite{yan2021sparse} & L & 30.2 & 12.5 & 14.2 & 3.4 & 13.6 & 12.0 & 7.2 & 4.3 & 7.3 & 6.8 & 9.2 & 9.1 & 27.9 & 15.3 & 14.9 & 16.2 & 14.0 & 24.9 \\
Co-Occ~\cite{gan2024coocc} & C\&L & 30.6 & 21.9 & 26.5 & 16.8 & 22.3 & 27.0 & 10.1 & 20.9 & 20.7 & 14.5 & 16.4 & 21.6 & 36.9 & 23.5 & 5.5 & 23.7 & 20.5 & 23.5 \\

\midrule

M-CONet \cite{zhang2024mconet} & C\&L & 29.5 & 20.1 & 23.3 & 13.3 & 21.2 & 24.3 & 15.3 & 15.9 & 18.0 & 13.3 & 15.3 & 20.7 & 33.2 & 21.0 & 22.5 & 21.5 & 19.6 & 19.8 \\
\rowcolor{myLightGray}
M-CONet \textbf{+ Ours} & C\&L & 32.1 & 22.4 & 26.8 & 15.1 & 23.2 & 25.9 & 18.4 & 19.2 & 20.5 & 15.7 & 17.8 & 23.9 & 35.1 & 23.6 & 24.9 & 24.1 & 21.2 & 22.3 \\
\midrule
OccMamba \cite{li2025occmamba} & C\&L & \underline{34.7} & \underline{25.2} & \underline{29.1} & \underline{19.1} & \underline{25.5} & \underline{28.5} & \underline{18.1} & \underline{24.7} & \underline{23.4} & \underline{19.8} & \underline{19.3} & \underline{24.5} & \underline{37.0} & \underline{25.4} & \underline{25.4} & \underline{25.4} & \underline{28.1} & \underline{24.9} \\
\rowcolor{myLightGray}

OccMamba \textbf{+ Ours} & C\&L & \textbf{37.0} & \textbf{26.6} & \textbf{32.9} & \textbf{20.7} & \textbf{27.4} & \textbf{29.6} & \textbf{22.0} & \textbf{28.4} & \textbf{24.6} & \textbf{21.8} & \textbf{21.1} & \textbf{27.0} & \textbf{39.3} & \textbf{27.9} & \textbf{27.6} & \textbf{27.4} & \textbf{29.9} & \textbf{26.6} \\
\bottomrule
\end{tabular}
\end{threeparttable}
}
\label{table_openoccupancy}
\end{table*}

\subsection{Depth-Aware Geometric Alignment Loss}



The DAGA loss is calculated following the Lift-Splat-Shoot (LSS)-based transformation to resolve structural discrepancies between $V_{cam}$ and $V_{pts}$. While LSS-based methods successfully map 2D image features into a 3D frustum, they inherently suffer from depth distribution ambiguity due to the ill-posed nature of monocular depth estimation. This often results in pseudo-point artifacts and significant vertical streaking, where features smear along the depth axis rather than localizing at specific 3D coordinates. DAGA is specifically designed to isolate and align the geometrical features by comparing the $L_2$-normalized, sigmoid-activated intensities of both voxel volumes, providing a structural supervisor that guides the camera-to-voxel mapping toward the geometric accuracy of the LiDAR reference.

To combat the bleeding effect across the $z$-axis, we introduce a sharpness constraint, $L_{sharp}$, which penalizes vertical gradient discrepancies:
\begin{equation}
    L_{sharp} = \| \nabla_z V_{cam} - \nabla_z V_{pts} \|_1
\end{equation}
where $\nabla_z$ denotes the first-order difference along the depth dimension. Furthermore, to prioritize near-field consistency where camera depth estimation is most reliable, we utilize a depth-dependent weighting function:
\begin{equation}
    W(d) = \frac{1}{1 + \beta  (d/D)}
\end{equation}
where $d \in \{0, \dots, D-1\}$ represents the iterative discrete depth index, $D=10$ is the total depth of the voxel grid, and $\beta$ is a decay hyperparameter. 

The total DAGA loss is calculated by iterating over the depth layers:
\begin{equation}
    L_{DAGA} = \frac{1}{D} \sum_{d=0}^{D-1} \left( \text{MSE}(V_{cam}^{(d)}, V_{pts}^{(d)}) W(d) \right) + \lambda L_{sharp}
\end{equation}
where $V^{(d)}$ represents the 2D feature slice at depth $d$ and $\lambda$ is a balancing coefficient set to 0.1. By penalizing vertical gradient differences and weighting by depth, $L_{DAGA}$ ensures that object boundaries remain distinct and well-defined.

\subsection{Objective function}
The total training objective $L_{total}$ is formulated as a multi-task loss that simultaneously optimizes semantic segmentation and geometric consistency. For semantic supervision, we employ a combination of voxel-wise cross-entropy loss $L_{ce}$ and Lov\'{a}sz-softmax loss \cite{berman2018lovasz} $L_{lov}$  to address class imbalance and directly optimize the mIoU metric. To ensure structural alignment, we integrate our proposed Depth-Aware Geometric Alignment loss $L_{DAGA}$. The joint objective is defined as:
\begin{equation}
L_{total} = L_{ce}(\hat{O}, \bar{O}) + L_{lov}(\hat{O}, \bar{O}) + \lambda_{daga} L_{DAGA}
\end{equation}
where $\hat{O}$ and $\bar{O}$ represent the predicted and ground-truth occupancy volumes, respectively. We set the loss coefficients to  $\lambda_{daga} = 0.2$ to balance semantic accuracy with geometric refinement.

\section{Experiments}

\subsection{Dataset}
We evaluate our framework on nuScenes \cite{caesar2020nuscenes} for multimodal 3D semantic occupancy and SemanticKITTI \cite{behley2019semantickitti} for semantic scene completion. For nuScenes \cite{caesar2020nuscenes}, we utilize the dense annotations from OpenOccupancy \cite{wang2023openoccupancy} across 1,000 synchronized sequences. Our occupancy grid is defined within a range of $[-50, 50]$m for $X, Y$ and $[-5, 3]$m for $Z$, yielding a $200 \times 200 \times 16$ resolution from $1600 \times 900$ pixel inputs. Following standard protocols, we train on the training split and report performance on the validation set. For SemanticKITTI \cite{caesar2020nuscenes}, we demonstrate generalizability using only the left RGB camera ($1241 \times 376$ resolution). The ground truth consists of 21 semantic categories voxelized into a $256 \times 256 \times 32$ grid with a 0.2m voxel size.

\subsection{Implementation Details}
To demonstrate the versatility and plug-and-play nature of our approach, we integrate the proposed InstVLM, WeathFusion, and DAGA modules into two distinct baseline architectures: OccMamba and MCoNet. Input images are processed at a resolution of $1600 \times 900$ for nuScenes and $1920 \times 1200$ for SemanticKITTI using a ResNet101 backbone  and an FPN neck. For the LiDAR branch, we aggregate the ten most recent point cloud sweeps to ensure temporal density before extracting features through a voxel encoder. The InstVLM module utilizes a CLIP-ViT-B/16 text encoder specialized via Low-Rank Adaptation (LoRA) to minimize trainable parameter overhead. The network is optimized via the AdamW optimizer with a weight decay of $0.01$ and a batch size of 1. We employ initial learning rates of $1 \times 10^{-4}$ for nuScenes and $2 \times 10^{-4}$ for SemanticKITTI, both of which follow a cosine annealing schedule over 20 training epochs. All experiments are conducted on 4x NVIDIA RTX 6000 ADA GPUs.

\subsection{Performance Evaluation Metrics}
To evaluate the efficacy of our framework against state-of-the-art (SOTA) methodologies, we utilize the Intersection over Union (IoU) to quantify the accuracy of each semantic class. Furthermore, we employ the mean IoU (mIoU) across all semantic categories to provide a holistic assessment of spatial-semantic fidelity, defined as follows:
\begin{equation}
    IoU=\frac{TP}{TP+FP+FN}, 
\end{equation}
and 
\begin{equation}
    \quad mIoU=\frac{1}{Cls}\sum_{i=1}^{Cls}  \frac{TP_{i}}{TP_{i}+FP_{i}+FN_{i}} 
\end{equation}
where $TP$, $FP$, and $FN$ denote the counts of true positives, false positives, and false negatives, respectively, and $Cls$ represents the total number of semantic classes.

\begin{figure*}[!ht]
  \centering
  \includegraphics[width=0.8\textwidth]{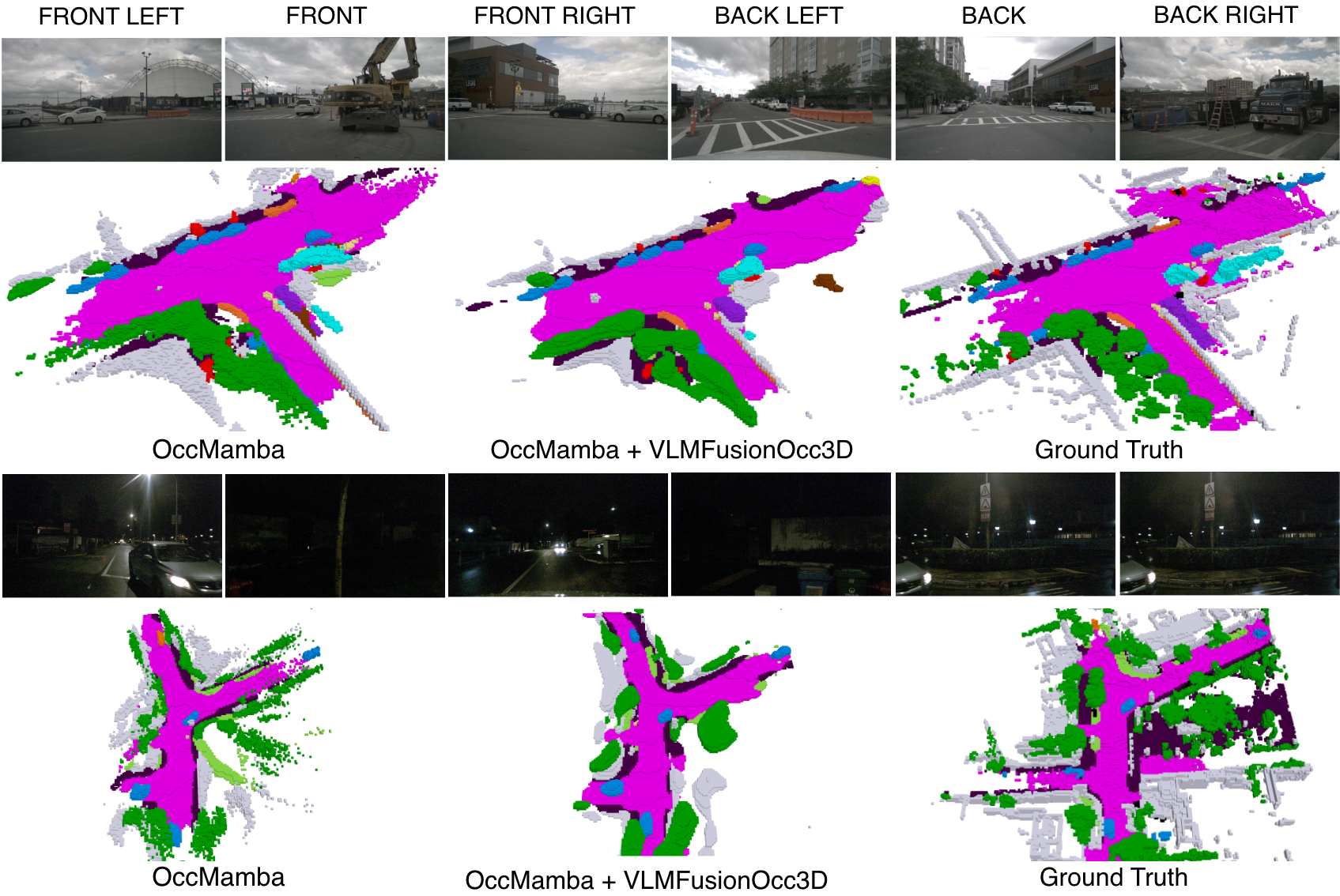}
  \vspace{-0.3cm}

  \caption{Qualitative results under adverse weather and lighting conditions. We compare the baseline OccMamba (middle) with our VLMFusionOcc3D (middle) across rainy and nighttime scenes from the nuScenes-OpenOccupancy dataset. Better viewed when zoomed in.}
  \label{fig:qual}
\end{figure*}

\begin{table}[htbp]
\centering
\caption{Performance on SemanticKITTI test set. The best and second-best are in bold and underlined, respectively.}
\vspace{-0.3cm}
\label{table_semantickitti}
\fontsize{9pt}{11pt}\selectfont
\begin{threeparttable}
\begin{tabular}{l|c|c} 
\toprule
Method & Input Modality & mIoU \\

\midrule

MonoScene~\cite{cao2022monoscene} & C & 11.1 \\
SurroundOcc~\cite{wei2023surroundocc} & C & 11.9 \\
OccFormer~\cite{zhang2023occformer} & C & 12.3 \\
RenderOcc~\cite{pan2023renderocc} & C & 12.8 \\
LMSCNet~\cite{roldao2020lmscnet} & L & 17.0 \\
JS3C-Net~\cite{yan2021sparse} & L & 23.8 \\
SSC-RS~\cite{li2024sscrs} & L & 24.2 \\
Co-Occ~\cite{gan2024coocc} & C\&L & \underline{24.4} \\
\midrule
M-CONet~\cite{zhang2024mconet} & C\&L & 20.4  \\
\rowcolor{myLightGray}

M-CONet~\cite{zhang2024mconet} \textbf{+ Ours} & C\&L & 22.6  \\
\midrule
OccMamba~\cite{li2025occmamba} & C\&L & \underline{24.6}\\
\rowcolor{myLightGray}

OccMamba~\cite{li2025occmamba} \textbf{+ Ours}  & C\&L & \textbf{26.4}\\

\bottomrule
\end{tabular}
\end{threeparttable}
\end{table}
\vspace{-0.3cm}

\subsection{Results and Discussions}
We evaluate VLMFusionOcc3D against state-of-the-art methods on the nuScenes-OpenOccupancy and SemanticKITTI benchmarks. To demonstrate the architectural agnosticism and plug-and-play versatility of our framework, we implement our proposed modules on two distinct voxel-based architectures: M-CONet \cite{zhang2024mconet} and OccMamba \cite{li2025occmamba}.

As detailed in Table \ref{table_openoccupancy}, our framework significantly enhances the performance of these voxel-based baselines across nearly all semantic categories. When integrated with OccMamba \cite{li2025occmamba}, our approach achieves a top-tier 37.0\% IoU and 26.6\% mIoU. A critical finding is the substantial boost in detecting vulnerable road users (VRU); for instance, the IoU for pedestrians and motorcycles increases to 24.6\% and 28.4\%, respectively. This improvement is attributed to the InstVLM module, which anchors sparse geometric voxels to stable linguistic concepts, effectively resolving the semantic ambiguity typically found in slender or complex-shaped objects.

Furthermore, the integration with M-CONet \cite{zhang2024mconet} showcases the framework's ability to refine established consistency-based models. By adding our modules, M-CONet \cite{zhang2024mconet} experiences a 2.3\% mIoU increase, reaching 22.4\%. This uplift is particularly prominent in class-specific results such as driveable surface and sidewalk, where our WeathFusion module optimizes sensor weighting to compensate for surface reflection noise. 

On the SemanticKITTI test set (Table \ref{table_semantickitti}), our method establishes a new state-of-the-art with 26.4\% mIoU, outperforming existing multimodal approaches like Co-Occ \cite{gan2024coocc}, M-CONet \cite{zhang2024mconet}, and OccMamba \cite{li2025occmamba}. The balanced performance across large-scale structures (e.g., manmade, vegetation) and small-scale dynamic objects suggests that weather-aware gating combined with VLM-driven attention provides a more comprehensive environmental representation than traditional geometric-only fusion.

Qualitatively, results in figure \ref{fig:qual} confirm the robustness of our framework in rainy and night conditions. While the baseline OccMamba \cite{li2025occmamba} suffers from structural noise and false negatives due to signal scattering or low-light contrast, OccMamba + Ours maintains high-fidelity occupancy boundaries. By utilizing WeathFusion to dynamically re-calibrate sensor trust based on real-time metadata, our model effectively suppresses sensor-specific artifacts, resulting in cleaner semantic silhouettes.

\begin{table}[htbp]
\centering
\caption{Ablation study of different components on the nuScenes-OpenOccupancy validation set with the OccMamba baseline.}
\vspace{-0.3cm}
\resizebox{0.75\linewidth}{!}{ 
\begin{tabular}{ccc|c|c}
\toprule
InstVLM & WeathFusion & DAGA &  IoU & mIoU \\ 
\midrule
      &            &               & 34.7 & 25.2\\ 
\checkmark &       &                & 35.6 &26.6\\ 
 & \checkmark &                  & 35.1 & 26.2\\ 
 &  &   \checkmark         &  34.9 &25.7\\
\checkmark & \checkmark &       & 36.6 &  26.0\\
\rowcolor{myLightGray}
\checkmark & \checkmark & \checkmark & 37.0 & 26.6  \\
\bottomrule
\end{tabular}
}
\label{table:abb}
\end{table}
\vspace{-0.3cm}

\subsection{Ablation Study}
To evaluate the individual contribution of each proposed component, we conduct an ablation study using the OccMamba \cite{li2025occmamba} baseline on the nuScenes-OpenOccupancy \cite{wang2023openoccupancy} validation set. As detailed in Table \ref{table:abb}, the integration of Instance-driven VLM Attention (InstVLM) provides the most substantial individual performance boost, elevating the mIoU from 25.2\% to 26.6\%. This gain highlights the effectiveness of using linguistic priors to resolve semantic ambiguities in complex 3D voxel spaces. Similarly, the Weather-Aware Adaptive Fusion (WeathFusion) module independently increases the mIoU to 26.2\%, demonstrating that dynamic, context-aware sensor re-weighting significantly improves perception reliability. The Depth-Aware Geometric Alignment (DAGA) loss further contributes by refining the structural consistency between modalities, resulting in a 25.7\% mIoU. When all three modules are combined, the framework achieves its peak geometric performance with a 37.0\% IoU and 26.6\% mIoU, confirming that the synergy between semantic grounding and environmental adaptation is crucial for robust 3D occupancy prediction.

\subsection{Feature Fusion Analysis}
Table \ref{table:fusion} evaluates different fusion strategies using the OccMamba baseline on an NVIDIA RTX 6000 Ada GPU. While elementary operations like Addition and Concatenation offer minimal latency, they lack the capacity to model complex environmental contexts. 3D convolutional layers offer a slight structural refinement but introduce a disproportionate increase in latency (0.65 ms) compared to simpler operations without significant mIoU gains. In contrast, our proposed WeathFusion achieves a superior 26.6\% mIoU, representing a 1.4\% improvement over standard Concatenation. Notably, WeathFusion outperforms the recent ACLF (GaussianOcc3D) \cite{doruk2026gaussianocc3d}, which utilizes a Gaussian-based fusion, by 1.1\% mIoU while reducing inference latency from 3.21 ms to 2.14 ms. These results indicate that utilizing VLM-based semantic anchors via gated cross-attention on the voxel grid provides a more effective trade-off between predictive accuracy and computational efficiency than traditional 3D convolutional or Gaussian-based fusion methods.

\begin{table}[htbp]
\centering
\caption{Performance analysis of different fusion techniques on nuScenes-OpenOccupancy  validation set  with the OccMamba baseline.}
\vspace{-0.3cm}
\resizebox{0.9\linewidth}{!}{ 
\begin{tabular}{l|c|c|c}
\toprule
Method &  IoU & mIoU & Latency (ms)\\ 
\midrule
 Addition    & 34.2 & 24.9 & 0.02 \\ 
 Concatenation & 34.7 & 25.2 & 0.03 \\ 
 3D Convolution   & 34.6 & 25.0 & 0.65 \\ 
 ACLF \cite{doruk2026gaussianocc3d}  &  35.6 & 25.5 & 3.21 \\
 \rowcolor{myLightGray}

 \textbf{WeathFusion (Ours)}   & 37.0 & 26.6 & 2.14 \\
\bottomrule
\end{tabular}
}
\label{table:fusion}
\end{table}

\subsection{Model Performance Analysis}
Table \ref{tab:efficiency} demonstrates the computational efficiency and versatility of our proposed module. When integrated with the M-CONet \cite{zhang2024mconet} and OccMamba \cite{li2025occmamba} backbones, our method yields significant performance gains of +2.3\% and +1.4\% mIoU, respectively. Notably, these improvements are achieved with minimal memory overhead, requiring only a 1.6 GiB increase in training memory and a negligible 0.5--0.6 GiB during inference. This efficiency is attributed to our use of a frozen CLIP text encoder \cite{li2023bevclip} and LoRA-based adaptation \cite{hu2023lora}, which leverages rich linguistic priors without the high cost of full VLM fine-tuning. 

\begin{table}[htbp]
\centering
\caption{Efficiency Analysis on nuScenes-OpenOccupancy Validation Set.}
\vspace{-0.3cm}
\label{tab:efficiency}
\resizebox{0.9\columnwidth}{!}{
\begin{tabular}{l|cc|c}
\hline
\multirow{2}{*}{Method} & \multicolumn{2}{c|}{Memory Usage (GiB)}  & mIoU \\
 & Training & Inference &  (\%) \\ \hline
M-CONet \cite{zhang2024mconet} & 37.3 & 12.1  & 20.1 \\
\rowcolor{myLightGray}
M-CONet \cite{zhang2024mconet}  \textbf{+ Ours} & 38.9 & 12.7 & 22.4 \\
OccMamba \cite{zhu2024occmamba} & 23.1 & 6.8  & 25.2 \\ 
\rowcolor{myLightGray}
OccMamba \cite{li2025occmamba} \textbf{+ Ours} & 24.7 & 7.3 & 26.6 \\ \hline
\end{tabular}
}
\end{table}
\vspace{-0.4cm}

\subsection{Adverse Condition Analysis}
To analyze the effectiveness of the proposed framework in challenging environments, we evaluate the performance of the OccMamba \cite{li2025occmamba} baseline and its integration with our modules under diverse weather and lighting conditions. As illustrated in Table \ref{tab:weather}, our approach yields significant performance gains across all scenarios, with the most notable improvements occurring in Rainy and Night conditions. Specifically, in rainy scenarios, the mIoU increases from 24.1\% to 29.3\%, an improvement of 5.2\%, which demonstrates the efficacy of WeathFusion in mitigating LiDAR signal degradation caused by precipitation. Even more striking is the performance during night-time operations, where the mIoU leaps from 11.8\% to 17.3\%. This 5.5\% boost highlights how the InstVLM module leverages robust semantic linguistic priors to compensate for the drastic contrast loss in camera-based features under low-light conditions. Even in the standard Day condition, our framework provides a solid improvement in IoU (from 36.8 to 40.2), indicating that the dynamic sensor weighting and depth alignment provided by our modules enhance geometric accuracy regardless of environmental degradation.

\begin{table}[htbp]
    \centering
    \caption{Performance comparison on the nuScenes-OpenOccupancy validation set under adverse conditions.}
    \vspace{-0.3cm}
    \resizebox{\linewidth}{!}{
     \begin{tabular}{l|c|ccc|ccc}
        \toprule
        & & \multicolumn{3}{c|}{IoU $\uparrow$} & \multicolumn{3}{c}{mIoU $\uparrow$} \\
        Method & Modality  & Rainy & Day & Night & Rainy & Day & Night \\ 
        \midrule
        OccMamba \cite{zhu2024occmamba} &  C\&L & 28.3 & 36.8 & 12.6 & 24.1 & 26.3 & 11.8 \\
        \rowcolor{myLightGray}
        OccMamba  \cite{zhu2024occmamba} \textbf{+ Ours} &  C\&L & 32.1 & 40.2 & 20.3 & 29.3 & 27.8 & 17.3 \\
        \bottomrule
      \end{tabular}
    }
    \label{tab:weather}
\end{table}

\section{conclusion}
In this paper, we presented VLMFusionOcc3D, a multimodal 3D semantic occupancy framework that effectively addresses semantic ambiguity and environmental sensitivity by integrating Vision-Language Models (VLMs) with weather-aware gating mechanisms. Our plug-and-play modules—InstVLM, WeathFusion, and the DAGA loss—enable dynamic re-weighting of sensor contributions based on environmental context derived from vehicle metadata, ensuring robust perception in adverse weather. Extensive evaluations on nuScenes and SemanticKITTI benchmarks demonstrate that our framework consistently improves the mIoU of voxel-based baselines, particularly for vulnerable road users. By anchoring sparse geometric features to stable linguistic embeddings, we provide a scalable solution that maintains high-fidelity spatial-semantic consistency across diverse operational domains.

{\small
\bibliographystyle{ieeetr}
\bibliography{ref}}

\begin{thebibliography}{10}

\bibitem{caesar2020nuscenes}
H.~Caesar {\em et~al.}, ``nuscenes: A multimodal dataset for autonomous driving,'' in {\em CVPR}, 2020.

\bibitem{behley2019semantickitti}
J.~Behley {\em et~al.}, ``Semantickitti: A dataset for semantic scene understanding of lidar sequences,'' in {\em ICCV}, 2019.

\bibitem{li2023bevformer}
Z.~Li {\em et~al.}, ``Bevformer: Learning bird's-eye-view representation from multi-camera images via spatiotemporal transformers,'' in {\em ECCV}, 2022.

\bibitem{huang2023triocc}
Y.~Huang {\em et~al.}, ``Tri-occ: Tri-perspective view for 3d semantic occupancy prediction,'' {\em arXiv preprint arXiv:2303.12345}, 2023.

\bibitem{yan2024inverse}
X.~Yan {\em et~al.}, ``Inversematrixvt3d: An efficient projection matrix-based approach for 3d semantic occupancy prediction,'' in {\em IROS}, 2024.

\bibitem{gan2024coocc}
Y.~Gan {\em et~al.}, ``Co-occ: Coupling explicit and implicit occupancy for 3d perception,'' {\em arXiv preprint arXiv:2403.01234}, 2024.

\bibitem{zhu2024occmamba}
X.~Zhu {\em et~al.}, ``Occmamba: Efficient 3d semantic occupancy prediction via state space models,'' {\em arXiv preprint arXiv:2404.12345}, 2024.

\bibitem{zhang2024mconet}
Y.~Zhang {\em et~al.}, ``Mconet: Multi-view consistency network for 3d semantic occupancy prediction,'' in {\em CVPR}, 2024.

\bibitem{zhou2024occfusion}
L.~Zhou {\em et~al.}, ``Occfusion: A multimodal fusion framework for 3d semantic occupancy prediction,'' {\em arXiv preprint arXiv:2401.09876}, 2024.

\bibitem{philion2020lift}
J.~Philion and S.~Fidler, ``Lift, splat, shoot: Encoding images from arbitrary camera rigs by anticipating video,'' in {\em ECCV}, 2020.

\bibitem{radford2021learning}
A.~Radford {\em et~al.}, ``Learning transferable visual models from natural language supervision,'' in {\em ICML}, 2021.

\bibitem{wang2023bevclip}
J.~Wang {\em et~al.}, ``Bev-clip: Multi-modal bev retrieval and forecasting with vision-language models,'' in {\em CVPR}, 2023.

\bibitem{hu2023lora}
E.~J. Hu {\em et~al.}, ``Lora: Low-rank adaptation of large language models,'' {\em arXiv preprint arXiv:2106.09685}, 2021.

\bibitem{huang2025gaussianformer}
Y.~Huang, A.~Thammatadatrakoon, W.~Zheng, Y.~Zhang, D.~Du, and J.~Lu, ``Gaussianformer-2: Probabilistic gaussian superposition for efficient 3d occupancy prediction,'' in {\em Proceedings of the computer vision and pattern recognition conference}, pp.~27477--27486, 2025.

\bibitem{doruk2026gaussianocc3d}
A.~Doruk and H.~F. Ates, ``Gaussianocc3d: A gaussian-based adaptive multi-modal 3d occupancy prediction,'' {\em arXiv preprint arXiv:2601.22729}, 2026.

\bibitem{chen2025m3net}
X.~Chen, S.~Shi, T.~Ma, J.~Zhou, S.~See, K.~C. Cheung, and H.~Li, ``M3net: Multimodal multi-task learning for 3d detection, segmentation, and occupancy prediction in autonomous driving,'' in {\em Proceedings of the AAAI Conference on Artificial Intelligence}, vol.~39, pp.~2275--2283, 2025.

\bibitem{li2023bevclip}
J.~Li {\em et~al.}, ``Bev-clip: Multi-modal bird's eye view learning for autonomous driving,'' {\em arXiv preprint arXiv:2308.11822}, 2023.

\bibitem{vobecky2023pop3d}
A.~Vobecky {\em et~al.}, ``Pop-3d: Open-vocabulary 3d occupancy prediction from images,'' in {\em proceedings of the Neural Information Processing Systems (NeurIPS)}, 2023.

\bibitem{yu2024locc}
Z.~Yu, B.~Pang, {\em et~al.}, ``Language driven occupancy prediction,'' in {\em Proceedings of the IEEE/CVF International Conference on Computer Vision (ICCV)}, 2024.

\bibitem{sun2023vlmocc}
H.~Sun {\em et~al.}, ``Vlm-occ: A zero-shot 3d occupancy prediction framework,'' {\em arXiv preprint arXiv:2311.16104}, 2023.

\bibitem{wu2024promptocc}
Y.~Wu {\em et~al.}, ``Promptocc: Dreaming of virtual tokens for open-vocabulary 3d occupancy prediction,'' in {\em proceedings of the Neural Information Processing Systems (NeurIPS)}, 2024.

\bibitem{cao2022monoscene}
A.-Q. Cao, R.~de~Charette, and Y.-Z. Chen, ``Monoscene: Monocular 3d semantic scene completion,'' in {\em Proceedings of the IEEE/CVF Conference on Computer Vision and Pattern Recognition}, pp.~3991--4001, 2022.

\bibitem{huang2023tri}
Y.~Huang, W.~Zheng, Y.~Zhang, J.~Zhou, and J.~Lu, ``Tri-perspective view for vision-based 3d semantic occupancy prediction,'' in {\em Proceedings of the IEEE/CVF Conference on Computer Vision and Pattern Recognition}, pp.~9223--9232, 2023.

\bibitem{tang2024sparseocc}
P.~Tang, Z.~Wang, G.~Wang, J.~Zheng, X.~Ren, B.~Feng, and C.~Ma, ``Sparseocc: Rethinking sparse latent representation for vision-based semantic occupancy prediction,'' in {\em Proceedings of the IEEE/CVF Conference on Computer Vision and Pattern Recognition}, pp.~15923--15933, 2024.

\bibitem{chen20203d}
X.~Chen {\em et~al.}, ``3d-sketch: Sketch-based 3d shape retrieval and reconstruction,'' in {\em Proceedings of the AAAI Conference on Artificial Intelligence}, 2020.

\bibitem{li2020anisotropic}
J.~Li, K.~Liu, J.~Wang, Y.-Z. Chen, {\em et~al.}, ``Anisotropic convolutional networks for 3d semantic scene completion,'' in {\em Proceedings of the IEEE/CVF Conference on Computer Vision and Pattern Recognition}, pp.~3351--3359, 2020.

\bibitem{roldao2020lmscnet}
L.~Roldao, R.~de~Charette, and A.~Verroust-Blondet, ``Lmscnet: Lightweight multiscale 3d semantic completion,'' {\em IEEE Robotics and Automation Letters}, vol.~5, no.~4, pp.~6456--6463, 2020.

\bibitem{yan2021sparse}
X.~Yan, J.~Gao, J.~Li, R.~Zhang, Z.~Li, R.~Huang, and S.~Cui, ``Sparse single sweep lidar point cloud segmentation via learning contextual shape priors from scene completion,'' in {\em Proceedings of the AAAI conference on artificial intelligence}, vol.~35, pp.~3101--3109, 2021.

\bibitem{li2025occmamba}
J.~Li {\em et~al.}, ``Occmamba: Semantic occupancy prediction with state space models,'' in {\em Proceedings of the IEEE/CVF Conference on Computer Vision and Pattern Recognition}, 2025.

\bibitem{berman2018lovasz}
M.~Berman, A.~R. Triki, and M.~B. Blaschko, ``The lov{\'a}sz-softmax loss: A tractable surrogate for the optimization of the intersection-over-union measure in neural networks,'' in {\em Proceedings of the IEEE conference on computer vision and pattern recognition}, pp.~4413--4421, 2018.

\bibitem{wang2023openoccupancy}
X.~Wang, Z.~Zhu, W.~Xu, Y.~Zhang, Y.~Wei, X.~Chi, Y.~Ye, D.~Du, J.~Lu, and X.~Wang, ``Openoccupancy: A large scale benchmark for surrounding semantic occupancy perception,'' in {\em Proceedings of the IEEE/CVF International Conference on Computer Vision}, pp.~17850--17859, 2023.

\bibitem{wei2023surroundocc}
Y.~Wei, L.~Liang, P.~Zhang, J.~Li, {\em et~al.}, ``Surroundocc: Multi-view 3d semantic occupancy prediction with spatiotemporal fusion,'' in {\em Proceedings of the IEEE/CVF International Conference on Computer Vision (ICCV)}, 2023.

\bibitem{zhang2023occformer}
Y.~Zhang, X.~Zhu, Y.~Hou, H.~Li, {\em et~al.}, ``Occformer: Dual-path transformer for 3d semantic occupancy prediction,'' in {\em Proceedings of the IEEE/CVF International Conference on Computer Vision (ICCV)}, 2023.

\bibitem{pan2023renderocc}
M.~Pan, J.~Liu, R.~Zhang, P.~Huang, X.~Li, H.~Xie, B.~Wang, L.~Liu, and S.~Zhang, ``Renderocc: Vision-centric 3d occupancy prediction with 2d rendering supervision,'' in {\em 2024 IEEE International Conference on Robotics and Automation (ICRA)}, pp.~12404--12411, IEEE, 2024.

\bibitem{li2024sscrs}
J.~Mei, Y.~Yang, M.~Wang, T.~Huang, X.~Yang, and Y.~Liu, ``Ssc-rs: Elevate lidar semantic scene completion with representation separation and bev fusion,'' in {\em 2023 IEEE/RSJ International Conference on Intelligent Robots and Systems (IROS)}, pp.~1--8, IEEE, 2023.

\end{thebibliography}

\end{document}